\documentclass[conference]{IEEEtran}
\ifCLASSINFOpdf
\else
\fi
\usepackage{amsmath}
\usepackage{graphicx}
\usepackage[utf8]{inputenc}
\usepackage{amssymb}
\usepackage{eurosym}
\usepackage{float}
\usepackage{color}
\usepackage{colortbl}
\usepackage{cancel}
\usepackage{algpseudocode}
\usepackage{algorithm}
\usepackage{url}

\newtheorem{definition}{Definition}

\newcommand{\mcrel}{|\hspace*{-0.5em}<}
\newcommand{\mncrel}{\cancel{\mcrel}}

\begin{document}
%
\title{Using temporal abduction for biosignal interpretation: A case study on
QRS detection}

\author{\IEEEauthorblockN{T.~Teijeiro, P.~F\'elix and J.~Presedo}
\IEEEauthorblockA{Centro de Investigaci\'on en
Tecnolox\'ias da Informaci\'on (CITIUS)\\
University of Santiago de Compostela\\
Santiago de Compostela, SPAIN 15782\\
Email: tomas.teijeiro@usc.es}}


%


\maketitle

\begin{abstract}
In this work, we propose an abductive framework for biosignal interpretation,
based on the concept of \textit{Temporal Abstraction Patterns}. A temporal
abstraction pattern defines an abstraction relation between an observation
hypothesis and a set of observations constituting its evidence support. New
observations are generated abductively from any subset of the evidence of a
pattern, building an abstraction hierarchy of observations in which higher
levels contain those observations with greater interpretative value of the
physiological processes underlying a given signal. Non-monotonic reasoning
techniques have been applied to this model in order to find the best
interpretation of a set of initial observations, permitting even to correct
these observations by removing, adding or modifying them in order to make them
consistent with the available domain knowledge. Some preliminary experiments
have been conducted to apply this framework to a well known and bounded problem:
the QRS detection on ECG signals. The objective is not to provide a new better
QRS detector, but to test the validity of an abductive paradigm. These
experiments show that a knowledge base comprising just a few very simple rhythm
abstraction patterns can enhance the results of a state of the art algorithm by
significantly improving its detection F1-score, besides proving the ability of
the abductive framework to correct both sensitivity and specificity failures.
\end{abstract}



%
\IEEEpeerreviewmaketitle

\section{Introduction}
A prominent objective of biosignal processing research is to provide
classification algorithms for identifying the interesting phenomena from signal
samples. Such classifiers can be grouped into two main approaches: (1) a
\textit{knowledge-based} approach, which aims to model the domain or,
alternatively, to model an expert reasoning process~\cite{Kundu00}; and a (2)
\textit{learning-based} approach, which builds a model by estimating the
underlying mechanisms that produce the data of a training set~\cite{Clifford06}.
Once the classifier is obtained and validated, it behaves at a logical level as
a deductive system from a data vector representing the signal. The present work
departs from the intrinsic limitations of the deductive framework for coping
with the interpretation of biosignals. Indeed, deductive approaches apply a
monotonic consequence relation, so that any conclusion cannot be retracted as
new evidence becomes available. This entails a propagation of errors from the
first processing stages onwards, narrowing the capability of making a proper
identification as new processing stages are successively added. Usually, both
the above-mentioned approaches overcome this weakness through an artificial
adoption of a casuistry-based strategy, yielding to unsatisfactory results so
far.

The present work starts from the hypothesis that temporal abductive
reasoning~\cite{Peng90} provides a more appropriate framework for the
computational interpretation of biosignals. Its greatest strength lies in its
non-monotonic nature, so that the conclusions are guessed as conjectures
inferred from the available evidence at each time, and further information, at
different levels of abstraction, can modify these conclusions. For example, a
deductive arrhythmia classifier cannot correctly identify an ECG fragment in
which the beats were not properly detected. In contrast, an abductive
interpreter could conjecture the presence and morphology of a beat from its
context, like a human can reconstruct a speech despite failing to identify all
its constituent sounds.

This paper provides an abductive framework for biosignal interpretation on the
basis of the notion of Temporal Abstraction Pattern, which is a knowledge
representation formalism that defines an abstraction relation among a set of
domain observables, structuring the problem domain in an abstraction hierarchy,
where higher abstraction levels convey more semantic content and more
interpretative results. Patterns are generated by means of attributed regular
grammars~\cite{Aho86}, extended in order to address the representation of
temporal knowledge through the Simple Temporal Problem
formalism~\cite{Dechter91}, previously adopted for temporal abduction in
diagnosis~\cite{Brusoni97}.

The rest of this paper is outlined as follows: Section~\ref{sec:definitions}
defines the main notions of the proposal. Section~\ref{sec:algorithm} describes
an interpretation algorithm, based on heuristic searching.
Section~\ref{sec:qrsdetection} shows the representation of basic ECG knowledge,
and how it is used for QRS detection. Finally, section~\ref{sec:validation}
discusses validation results with respect to a state of the art algorithm, and
section~\ref{sec:conclusion} provides some conclusions.

\section{Definitions}
\label{sec:definitions}
A Simple Temporal Problem (STP)~\cite{Dechter91} defines a set of temporal
constraints between pairs of temporal variables, $T_i$ and $T_j$. A temporal
constraint is a closed interval $L(T_i,T_j) = [a_{ij}, b_{ij}]$ restricting the
admissible values for the difference $T_j-T_i$, as $a_{ij}\leq T_j-T_i\leq
b_{ij}$. Formally, a STP can be represented as a tuple
$N=\langle\mathcal{T},\mathcal{L}\rangle$, where
$\mathcal{T}=\{T_1,\ldots,T_n\}$ is a set of temporal variables, and
$\mathcal{L}=\{L(T_i,T_j);1\leq i,j\leq n\}$ is a set of temporal constraints
between them. A tuple $(t_1,...,t_n)$ is called a solution of $N$ if the
assignment $T_1=t_1$,...,$T_n=t_n$ satisfies all the constraints.

The basic representation entity of our abductive framework is the
\textbf{observable}, which is formally defined as a tuple $q=\langle
\eta,\vec{A},T_b, T_e \rangle$, where $\eta$ is the name of the observable,
$\vec{A}=(A_1,...,A_{n_q})$ is a set of attributes to be valued, and $T_b$ and
$T_e$ are two temporal variables representing the beginning and the end of the
observable. We denote by $\mathcal{Q}=\{q_0,q_1,...,q_n\}$ the set of
observables of a particular domain, comprising the vocabulary to describe the
phenomena of interest. An observable can be observed in multiple instances
called \textbf{observations}. An observation is defined as a tuple $o=\langle
\eta,\vec{v},t_b, t_e\rangle$, where $\eta$ is the name of the observable being
instantiated, $\vec{v}$ is a value assignment for each attribute of the
observable, and $t_b$ and $t_e$ are the specific values for the temporal
variables $T_b$ and $T_e$. We denote by $O(q)=\{o^q_1,...,o^q_i,...\}$ the set
of observations of the observable $q$. An example of an observable is $q=\langle
\mathtt{QRS}, \mathtt{shape}, T_b, T_e\rangle$, which represents a QRS complex
with a single attribute describing its morphology. This observable may be
instanced in different observations, like for example $o^q_1 =
\langle\mathtt{QRS}, \mathtt{shape=QS}, \text{00:32.123}, \text{00:32.201}
\rangle$.

Observables are related through abstraction relations defined by temporal
abstraction patterns. A temporal abstraction pattern models the necessary
knowledge to allow the conjecture of an observation of a high-level observable
from observations of lower level observables. We provide a procedure for
dynamically generating abstraction patterns, based on the formal language
theory. The set $\mathcal{Q}$ of observables can be considered as an alphabet
represented by their corresponding names. Given an alphabet $\mathcal{Q}$, a
formal grammar $G$ denotes a pattern of symbols of the alphabet, describing a
language $L(G)\subseteq \mathcal{Q}^\ast$, as a subset of the set of possible
strings of symbols of the alphabet. Let $G^{ap}$ be the class of formal
attributed grammars of abstraction patterns. A grammar $G\in G^{ap}$ is
syntactically defined as a tuple $(V_N, V_T, H, R)$. The production rules in $R$
are of one of the following forms: 
\begin{align} 
H=q_H &\rightarrow q [l] C \nonumber\\ 
    C &\rightarrow q[l]D~|~q[l]~|~\lambda \nonumber 
\end{align} 
$H=q_H$ is the initial symbol of the grammar, and it plays the role of the
hypothesis guessed by the pattern. $V_N$ is the set of non-terminal symbols of
the grammar. $V_T$ is the set of terminal symbols of the grammar, gathering
together: a set of observables $Q_G\subseteq \mathcal{Q}$, being $q\neq q_H$ for
all $q\in Q_G$, that can be abstracted by the hypothesis; a set of temporal
descriptions $[l]$ in the form of conjunctions of constraints between the
temporal variables of the observable produced by the rule and all the
observables previously generated; and the empty string, that is represented by
$\lambda$. Some simple examples of grammars to describe different basic cardiac
rhythms in terms of QRS complexes are detailed in
section~\ref{sec:qrsdetection}.

Given a grammar $G\in G^{ap}$, we provide a constructive method for representing
a set of abstraction patterns $P_G=\{P_1,...,P_i,...\}$. An abstraction pattern
shows a temporal arrangement between a set of observables, possibly appearing
repeatedly, to be abstracted by a new observable $q_H$. We call
\textbf{findings} to these occurrences of observables that lead to $q_H$, which
informally behave as observations that have not yet been observed, that is,
predictions generated by the grammar defining an abstraction pattern. Thus,
$M^q_P=\{m^q_1,m^q_2,...,m^q_i=\langle \eta,\vec{A},T^i_b, T^i_e\rangle,...\}$
is the set of findings of the observable $q$ in $P$.  $P_G$ gathers together the
set of abstraction patterns that share the same observable $q_H$ to be
abstracted, so they represent the different ways of hypothesizing $q_H$.

\begin{definition}
Given $G\in G^{ap}$, a \textbf{temporal abstraction pattern} $P=\langle
q_{H_P},M_P,N_P,\Pi_P\rangle$ consists of a hypothesis $q_{H_P}$; a set of
findings $M_P = \bigcup_{q \in \mathcal{Q}_G} M^q_P$ that form the evidence
supporting $q_{H_P}$; a temporal network $N_P$ between the temporal variables
involved in $q_{H_P}$ and $M_P$, defined by the $[l]$ attributes of $G$; and an
observation procedure $\Pi_P$ to compute the attribute values of $q_{H_P}$ from
the observed evidence.
\end{definition}

Below we show how an abstraction pattern is built by following the productions
of a grammar. In every step, a new observable is added, and a set of temporal
constraints among this finding and those generated before is introduced in
$N_P$.
\par
\noindent 1) Symbol $H$ entails initializing an abstraction pattern:
\begingroup\makeatletter\def\f@size{9}\check@mathfonts
\begin{displaymath}
 P\leftarrow \langle q_{H_P}, M_P=\varnothing , \langle
\mathcal{T}_P=\{T^H_b,T^H_e\},\mathcal{L}_P=\{L(T^H_b,T^H_e)\}\rangle \rangle
\end{displaymath}
\endgroup
\noindent  2) All those productions $H=q_{H_P} \rightarrow q[l]C$ entail:
\begingroup\makeatletter\def\f@size{9}\check@mathfonts
\begin{displaymath}
 P\leftarrow \langle q_{H_P}, M_P=  \{m_1^q\}, \langle \mathcal{T}_P \cup
\{T^{1}_b,T^{1}_e\}, \mathcal{L}_P \cup \mathcal{L}(P,m_1^q)\rangle \rangle,
\end{displaymath}
\begin{displaymath}
\mathcal{L}(P,m_1^q)=L(T^1_b,T^1_e)\cup\{L(T_i,T^{1}_j)| T_i\in
\mathcal{T}_P\wedge T^{1}_ j\in \{T^{1}_b,T^{1}_e\}\}
\end{displaymath}
\endgroup
\noindent 3) All those productions $C \rightarrow q [l]D~|~q[l]$ entail:
\begingroup\makeatletter\def\f@size{9}\check@mathfonts
\begin{displaymath}
  P\leftarrow \langle q_{H_P}, M_P \cup  \{m_k^q\}, \langle \mathcal{T}_P \cup 
\{T^{k}_b,T^{k}_e\}, \mathcal{L}_P \cup \mathcal{L}(P,m_k^q)\rangle \rangle,
\end{displaymath}
\begin{displaymath}
 \mathcal{L}(P,m_k^q)=L(T^k_b,T^k_e)\cup\{L(T_i,T^{k}_j)| T_i\in
\mathcal{T}_P\wedge T^{k}_ j\in \{T^{k}_b,T^{k}_e\}\}
\end{displaymath}
\endgroup
In case a temporal descriptor is omitted in a production of the $G^{ap}$
grammar, it is assumed the 'after' relationship between the new manifestation
and the set of previous manifestations. For instance, all those productions
$C\rightarrow qD~|~q$  entail:
\begingroup\makeatletter\def\f@size{9}\check@mathfonts
\begin{displaymath}
P\leftarrow \langle q_H, M_P \cup  \{m_k^q\}, \langle \mathcal{T}_P \cup 
\{T^{k}_b,T^{k}_e\}, \mathcal{L}_P \cup \mathcal{L}(P,m_k^q)\rangle \rangle,
\end{displaymath}
\begin{displaymath}
 \mathcal{L}(P,m_k^q)=L(T^k_b,T^k_e)\cup\{L(T_i,T^{k}_b)\subseteq
\mathbb{Z}^+|  T_i\in \{T^j_b,T^j_e\}\wedge m_j\in M_P\}
\end{displaymath}
\endgroup

\begin{definition}
 Let $\mathcal{Q}$ be a set of observables and
$\mathcal{P}$ a set of abstraction patterns. We say
$\mathcal{P}$ induces an \textbf{abstraction relation} in $\mathcal{Q} \times
\mathcal{Q}$, denoted by $q_i \mcrel q_j$ if and only if there exists a pattern
$P \in \mathcal{P}$ such that:
\begin{enumerate}
 \item $q_i \in M_P$
 \item $q_j = q_{H_P}$
 \item $q_i \mncrel^+ q_i$, where $\mcrel^+$ is the transitive closure of
$\mcrel$
\end{enumerate}
\end{definition}

\begin{figure}
 \includegraphics[width=\columnwidth]{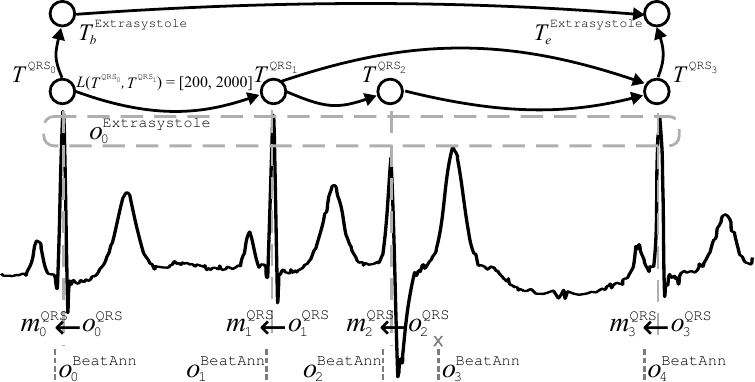}
 \caption{Example of an abstraction hypothesis of the \texttt{Extrasystole}
pattern. At the bottom of the figure the five initial beat annotations are
shown. Four of these observations are abstracted by \texttt{QRS} observations
that are matched with the evidence of the \texttt{Extrasystole} pattern,
allowing to discard the incorrect $o^\mathtt{BeatAnn}_3$ observation. The top of
the figure displays the constraint network of the pattern.}
 \label{fig:matchrel}
 \vspace{-12pt}
\end{figure}

The abstraction relation allows us to say that ``\textit{$q_j$ abstracts
$q_i$}''. This gives a hierarchy structure to the observables of a domain,
allowing us to define an abstraction model from a set of abstraction patterns.
Fig.~\ref{fig:matchrel} illustrates this hierarchy with the observables of the
knowledge base defined in section~\ref{sec:qrsdetection}.

\begin{definition}
 We define an \textbf{abstraction model} as a tuple $\mathcal{M} = \langle
\mathcal{P}, \mathcal{Q}, \mcrel \rangle$ where $\mathcal{P}$ induces an
abstraction relation $\mcrel$ over a set of domain observables $\mathcal{Q}$. 
\end{definition}

\begin{definition}
 We define an \textbf{interpretation problem} as a tuple $IP=\langle\mathcal{O},
\mathcal{M}\rangle$, where $\mathcal{O}= \{o_1,...,o_i,...\}$ is a set
of observations requiring interpretation, and $\mathcal{M}$ is an abstraction
model of the domain.
\end{definition}

An interpretation problem so defined gives a different status to the evidence
with respect to abductive diagnosis, where the explicit difference between
normal and faulty behaviors leads to the definition of faulty
findings~\cite{Poole89}. Only when a faulty finding is provided, the diagnostic
process is triggered. In contrast, an interpretation problem based on an
abstraction model gives all the findings the same status. The objective is to
provide a new description in terms of those observables with the highest
possible abstraction level. To do so, and under certain constraints, an
observation is tried to be assigned to a specific finding of an abstraction
pattern, so that a new observation is obtained for the hypothesis of the
pattern.

\begin{definition}
Given an interpretation problem $IP$, a \textbf{matching relation} for a pattern
$P\in \mathcal{P}$ is an injective relation in $ M_P \times \mathcal{O}$,
defined by  $m_i^q \leftarrow o_j$, iff $m_i^q=\langle \eta,\vec{A},T^i_b,
T^i_e\rangle \in M_P$ and $o_j=\langle \eta,\vec{v}_j,t^j_b, t^j_e\rangle \in
\mathcal{O}$.
\end{definition}

From the notion of matching relation we can design a mechanism for abductively
interpreting a subset of observations in $\mathcal{O}$ through the use of
abstraction patterns. Thus, a matching relation for a given pattern allows to
hypothesize new observations from previous ones, and to iteratively incorporate
new evidence to the interpretation by means of a hypothesis generation-and-test
cycle. The notion of abstraction hypothesis defines those conditions that a
subset of observations must satisfy in order to be abstracted by a new
observation.

\begin{definition}
 Given an interpretation problem $IP$, we define an \textbf{abstraction
hypothesis} as a tuple $h=\langle o^H_h,P_h,\leftarrow_h\rangle$, where
$P_h\in \mathcal{P}$, $\leftarrow_h \subseteq M_{P_h}\times \mathcal{O}$, and we
denote $O_h= \{o_{j1},\ldots,o_{jn}\} = image(\leftarrow_h)$, satisfying: 
\begin{enumerate}
\item $o^{H}_h\in O(q_{H_P})$
\vspace{0.2em}
\item $o^{H}_h=\Pi_P(O_h)$
\vspace{0.2em}
\item $(t^H_b, t^H_e, t^{j1}_b,...,t^{jn}_e)$ satisfy the constraints of $N_P$.
\end{enumerate}
\end{definition}

An abstraction hypothesis assigns a set of observations to the findings of the
pattern, giving them the role of evidence for the hypothesis. Even though the
matching relation is a matter of choice, and therefore a conjecture by itself,
some additional constraints could be assumed as default assumptions. An
important default assumption in the abstraction of periodic processes states
that consecutive observations are related by taking part of the same hypothesis,
defining the basic period of the process. This assumption leads to consecutive
findings of an observable within an abstraction pattern to be matched to
consecutive observations during the abstraction task.

As a result of an abstraction hypothesis, a new observation $o^H_h$ is
generated, that should be included in the set of domain observations, so
$\mathcal{O}=\mathcal{O} \cup \{o^H_h\}$. The abstraction process then iterates
until no new observations are generated. The set of observations that may be
abstracted in an interpretation problem $IP$ is $O(domain(\mcrel))$, that is,
the set of observations corresponding to observables playing the role of
findings in some abstraction pattern. Figure~\ref{fig:matchrel} shows an example
of an abstraction hypothesis of the \texttt{Extrasystole} pattern.

\begin{definition}
 Given an interpretation problem $IP$, we define an \textbf{interpretation} as a
temporally consistent set of abstraction hypothesis $I=\{h_1,...,h_m\}$,
satisfying $O_{h_i} \cap O_{h_j} = \emptyset$, $i\neq j$.
\end{definition}

The restriction $O_{h_i} \cap O_{h_j} = \emptyset$ establishes that in the same
interpretation an observation cannot be abstracted by more than one hypothesis,
formalizing the notion of alternative hypotheses as those abstracting the same
observation. An interpretation can be rewritten as $I=\langle
O^H_{I},P_I,\leftarrow_I,N_I\rangle$, where: $O^H_I=\{o^H_1,...,o^H_m\}$ is the
set of conjectured observations; $P_I=\{P_1,...,P_m\}$ is the set of abstraction
patterns used in the interpretation; $\leftarrow_I = \leftarrow_1 \cup ... \cup
\leftarrow_m \subseteq (M_{P_1}\cup ... \cup M_{P_m})\times \mathcal{O}$ is the
global matching relation;  and $N_I=N_1\cup ... \cup N_m$ is a network between
all the temporal variables involved. The temporal assignment made by
$\leftarrow_I$ is a solution of $N_I$.

\section{Interpretation algorithm}
\label{sec:algorithm}
The purpose of the interpretation algorithm is to obtain the best interpretation
for a given set of initial observations. Since a number of different sets of
abstraction hypotheses may be obtained with the same base evidence, we need a
method to discriminate some interpretations against other. To evaluate and
compare the generated interpretations, we rely on the following general
principles: The \textit{coverage principle}, which prefers interpretations
explaining more initial observations, and that is calculated as $C(I) =
|\bigcup_{h \in I} O_h|/|O(domain(\mcrel))|$; the \textit{simplicity principle},
also known as Occam's razor, which prefers interpretations with fewer
hypotheses, and that is calculated as $S(I) = 1/(1+|O^H_I|)$; the
\textit{abstraction principle}, which prefers interpretations using terms of
higher abstraction levels; and the \textit{predictive principle}, which prefers
interpretations that properly predict future observations.

Once we are able to evaluate and compare interpretations, solving an
interpretation problem can be posed as a heuristic search on the space of
consistent interpretations. The proposed search strategy uses coverage as the
main heuristic function, and considers simplicity when the coverage is equal in
more than one interpretation. The main drawback of coverage as heuristic is that
is non-admissible, and therefore optimality can not be guaranteed, so we require
an algorithm efficient with this type of heuristics and that saves memory and
computing time. We propose an evolution of the K-BFS method~\cite{Felner03},
named \textit{Partial Expansion K-BFS}, whose pseudocode is shown in
Algorithm~\ref{alg:pekbfs}. The algorithm takes as input an interpretation
problem $IP$, and returns the first found interpretation with full coverage, or
the interpretation with the highest coverage factor and highest simplicity if no
full coverages are found. The algorithm also needs a $K$ parameter, which
determines its exploratory capabilities, and which is set to the maximum number
of abstractions that can be made to an observable: $K=max(\{|\{q_j ~|~ q_i
\mcrel q_j\}|, q_i \in \mathcal{Q}\})$. The intuition behind this decision is
that at a certain point in the interpretation, and with equal coverage factor,
we give an opportunity to every possible hypothesis to continue the
interpretation.

The distinctive features of PE-KBFS with respect to KBFS are that (1) in each
node expansion, only one successor is obtained, greatly limiting the growth of
the \texttt{open} list; and (2) the heuristic estimation of the newly generated
nodes is propagated to the parent node, resulting in a significant reduction in
the number of expansions provided that child nodes are generated ordered by
their valuation. These two features are aimed at reducing the number of node
expansions, which in our case results in fewer generated interpretations.

\setlength{\textfloatsep}{10pt}
\begin{algorithm}[t]
\caption{Partial Expansion K-Best First Search algorithm.}
\label{alg:pekbfs}
\begin{algorithmic} [1]
\small
  \Function{PE-KBFS}{$IP, K$}
    \State \textbf{var} $I^0 = \langle
\emptyset,\emptyset,\emptyset,\emptyset\rangle$
    \State \Call{set\_focus}{$I^0, min_{t^i_b}(o_i \in \mathcal{O})$}
    \State \textbf{var} $open =$ \Call{sorted}{$[(\langle 1.0, 1.0\rangle,
I^0)]$}
    \State \textbf{var} $closed =$ \Call{sorted}{$[]$}
    \While{$open \neq \emptyset$}
      \ForAll{$I \in open[0\ldots K]$}
	\State $I' = $ \Call{next}{{\footnotesize GET\_SUCCESSORS}($I$)}
	\If{\textbf{not} $I'$}	  
	  \State $closed = closed \cup \{(\langle 1-C(I),
1/S(I)\rangle,I)\}$
	\ElsIf{$C(I') = 1.0$}
	  \State \Return $I'$
	\EndIf
	\State \textbf{var} $val = \langle 1-C(I'), 1/S(I')\rangle$
	\State $open = open \cup \{(val,I'),(val,I)\}$
      \EndFor      
    \EndWhile
    \State \Return $min(closed)$
  \EndFunction  
\end{algorithmic}
\end{algorithm}

The search process starts with the so-called trivial interpretation $I^0$, the
one with no abstraction hypotheses. At each step, each of the $K$ best
interpretations generated so far is evolved in a hypothesis-and-test cycle. This
cycle is driven by the concept of \textbf{focus of attention}, which is
established according to the predictive and abstraction principles and enables
to combine different inference modes based on the state of the interpretation.
The focus is initially set to the earliest initial observation in the
interpretation problem, and its evolution depends on the state of the
hypothesis-and-test cycle, managed by the
\begin{footnotesize}GET\_SUCCESSORS()\end{footnotesize} function, detailed in
Algorithm~\ref{alg:interpretation_succ}.

\begin{algorithm}[t!]
\caption{Method for obtaining the successors of an interpretation in the
adopted reasoning scheme.}
\label{alg:interpretation_succ}
\begin{algorithmic} [1]
\small
  \Function{get\_successors}{$I$}
    \State \textbf{var} $focus =$ \Call{get\_focus}{$I$}
    \State \textbf{var} $succ = \emptyset$
    \If{\Call{is\_observation}{$focus$}}
      \State $succ = $ \Call{abduce}{$I, focus$}
    \ElsIf{\Call{is\_finding}{$focus$}}
      \State $succ = $ \Call{subsume}{$I, focus$} $\cup$ \Call{deduce}{$I,
focus$}    
    \EndIf
    \State \Return $succ$    
  \EndFunction  
  \vspace{0.3em}
  \Function{subsume}{$I, f$}
    \State \textbf{var} $succ = \emptyset$
    \ForAll{$o_i \in \mathcal{O} ~|~  f \leftarrow o_i$}
      \State $I' = \langle O^H_I , P_I,  \leftarrow_I \cup ~f \leftarrow o_i,
								    N_I \rangle$
      \State $succ = succ \cup \{I'\}$
    \EndFor
    \State \Return $succ$
  \EndFunction
  \vspace{0.3em}
  \Function{deduce}{$I, f$}
    \State \textbf{var} $succ = \emptyset$
    \ForAll{$\{P \in \mathcal{P} ~|~ f \in O(q_{H_P})\}$}
      \State $h = \langle o^{H_P}_h, P, \emptyset \rangle$
      \State $I' = \langle O^H_I \cup \{o^{H_P}_h\}, P_I \cup \{P\},
\leftarrow_I \cup~f \leftarrow o^{H_P}_h, N_I \rangle$
      \State $\mathcal{O} = \mathcal{O} \cup \{o^{H_P}_h\}$      
      \ForAll{$m^q \in~$\Call{next\_findings}{$P$}}
	\State $I'' = \langle O^H_{I'}, P_{I'}, \leftarrow_{I'}, N_{I'}\rangle$
	\State \Call{set\_focus}{$I'', m^q$}
	\State $succ = succ \cup \{I''\}$
      \EndFor      
    \EndFor
    \State \Return $succ$
  \EndFunction
  \vspace{0.3em}
  \Function{abduce}{$I,o_i$}
    \State \textbf{var} $succ = \emptyset$
    \State \textbf{var} $\mathcal{P}^{o_i} = \{P \in \mathcal{P} ~|~ o_i
\in O(q_i), q_i \in M_P\}$
    \ForAll{$P \in \mathcal{P}^{o_i}$}
      \ForAll{$m^{q_i} \in M^{q_i}_P$}
	\State $h = \langle o^{H_P}_h, P, m^{q_i} \leftarrow o_i \rangle$
	\State $I' = \langle O^H_I \cup \{o^{H_P}_h\}, P_I \cup \{P\}, 
	\leftarrow_I \cup \leftarrow_h, N_I \rangle$
	\State $\mathcal{O} = \mathcal{O} \cup \{o^{H_P}_h\}$
	\State \Call{set\_focus}{$I', o^{H_P}_h$}
	\State $succ = succ \cup \{I'\}$
	\ForAll{$m^q \in~$\Call{next\_findings}{$P$}}
	  \State $I'' = \langle O^H_{I'}, P_{I'}, \leftarrow_{I'},
N_{I'}\rangle$
	  \State \Call{set\_focus}{$I'', m^q$}
	  \State $succ = succ \cup \{I''\}$
	\EndFor
      \EndFor      
    \EndFor
    \State \Return $succ$
  \EndFunction
\end{algorithmic}
\end{algorithm}

In the \textit{hypothesis} step, the focus is on an observation not yet
abstracted. Then, abductive reasoning is applied and a hypothesis of a higher
abstraction level is generated. This is implemented in the
\begin{footnotesize}ABDUCE()\end{footnotesize} function, which explores all the
patterns that may abstract the focused observation and, for each of them, it
transfers the focus to the evidence findings predicted by the pattern, switching
to the \textit{test} step of the cycle. This reasoning mode enforces the
abstraction principle, since it always generates a hypothesis of high
abstraction level on the basis of an observation of a lower level.

In the \textit{test} step, the focus is on a finding predicted by an abstraction
pattern of the current interpretation. In this case, two possible inference
modes may be used, \textit{subsumption} and \textit{deduction}. The
\begin{footnotesize}SUBSUME()\end{footnotesize} function looks for an existing
observation that can be consistently matched with the focus finding, while the
\begin{footnotesize}DEDUCE()\end{footnotesize} function creates a new
observation hypothesis for the matching and then looks for the lower level
evidence supporting it. Since an unmatched finding can be seen as a prediction
of the knowledge involved in an interpretation, these two inference modes 
enforce the predictive principle, but inasmuch as subsumption leads to simpler
interpretations it is preferred over deduction.

The \begin{footnotesize}NEXT\_FINDINGS()\end{footnotesize} function, used both
in abduction and deduction, allows to extend an existing hypothesis, returning
the possible next findings that may generate a pattern $P$, obtained through the
analysis of the $G_P$ grammar. If this function returns nothing, then the focus
of attention is set to the earliest observation in the domain of the abstraction
relation with the highest possible abstraction level, starting a new
hypothesis-and-test cycle.

\section{An application to ECG QRS detection}
\label{sec:qrsdetection}
In order to test the practical feasibility of this framework, we have defined a
simple knowledge base to tackle a well known and bounded problem: QRS detection
on ECG signals. This knowledge base assumes as initial observations the QRS
annotations identified by any other algorithm, and describes some of the most
common rhythm patterns in terms of temporal distances between beats. The
temporal consistency with any of these patterns allows us to discriminate those
annotations that do not really correspond with QRS complexes, as well as to
proactively search for those QRS that, being not initially annotated, are
predicted by any of the patterns. 

The knowledge base defines the following set of observables $\mathcal{Q} =
\{q_0, q_1, q_2, q_3, q_4, q_5\}$, with:
{\small%
\begin{align*}
 q_0 &= \langle \mathtt{BeatAnn}, \emptyset, T_b = T_e = T \rangle
 & q_1 &= \langle \mathtt{QRS}, \emptyset, T_b = T_e = T \rangle\\
 q_2 &= \langle \mathtt{NormalRhythm}, \emptyset, T_b, T_e \rangle
 & q_3 &= \langle \mathtt{Bradycardia}, \emptyset, T_b, T_e \rangle\\
 q_4 &= \langle \mathtt{Tachycardia}, \emptyset, T_b, T_e \rangle
 & q_5 &= \langle \mathtt{Extrasystole}, \emptyset, T_b, T_e \rangle
\end{align*}
}
$q_0$ represents a beat annotation, the starting point of the interpretation
process, and that is considered an instantaneous observable, hence $T_b=T_e$;
$q_1$ represents the QRS complex, which is also an instantaneous observable;
$q_2$, $q_3$ and $q_4$ are different observables for regular cardiac rhythms,
and the only difference between them is the maximum and minimum admitted
distance between consecutive beats; $q_5$ represents an extrasystole, which is a
cardiac rhythm characterized by the presence of a premature beat and the
subsequent return to the previous rhythm with a compensatory pause.

Below we show the grammars and observation procedures relating these observables
(all temporal constraints are expressed in milliseconds):
\begin{align*}
G_{P_0}:& H = \mathtt{BeatAnn} & \rightarrow & \lambda\\
G_{P_1}:& H = \mathtt{QRS} &\rightarrow & \mathtt{BeatAnn} [l_1]\\
G_{P_2}:& H = \mathtt{(N|B|T)Rhythm} & \rightarrow & \mathtt{QRS}_0 [l_{20}]~A\\
	& A & \rightarrow & \mathtt{QRS}_1[l_{21}]~B\\
	& B & \rightarrow & \mathtt{QRS}_n[l_{2n}]~B~|~\mathtt{QRS}_n [l_{2n}]\\
G_{P_5}:& H = \mathtt{Extrasystole} &\rightarrow & \mathtt{QRS}_0 [l_{50}]~C\\
        & C & \rightarrow & \mathtt{QRS}_1 [l_{51}]~D\\
        & D & \rightarrow & \mathtt{QRS}_2 [l_{52}]~E\\
        & E & \rightarrow & \mathtt{QRS}_3 [l_{53}]
\end{align*}
{\footnotesize
\begin{align*}
[l_1]    &= \{L(T^\mathtt{QRS}, T^\mathtt{BeatAnn}) = [-150,150]\} 	      \\
[l_{20}] &= \{L(T^{\mathtt{QRS}_0}, T^H_b) = [0,0]\}                          \\
[l_{21}] &= \{\mathtt{NRR} = [475,1333],\mathtt{BRR} = [1000, 2000],
						     \mathtt{TRR}=[200,600],  \\
	 &\qquad L(T^{\mathtt{QRS}_0},T^{\mathtt{QRS}_1})=\mathtt{(N|B|T)RR}\}\\
[l_{2n}] &= \{\mathtt{RR} = t^{\mathtt{QRS}_{n-1}}-t^{\mathtt{QRS}_{n-2}},    \\
         &\qquad L(T^{\mathtt{QRS}_{n-1}},T^{\mathtt{QRS}_n})= 
           \mathtt{(N|B|T)RR} \cap [0.5\cdot\mathtt{RR},1.5\cdot\mathtt{RR}], \\
	 &\qquad L(T^{\mathtt{QRS}_n},T^H_e) = [0,0]\}                        \\
[l_{50}] &= \{L(T^{\mathtt{QRS}_0}, T^\mathtt{Extrasystole}_b) = [0,0]\}      \\
[l_{51}] &= \{L(T^{\mathtt{QRS}_0}, T^{\mathtt{QRS}_1}) = [200, 2000]\}       \\
[l_{52}] &= \{\mathtt{RR0} = t^{\mathtt{QRS}_1}-t^{\mathtt{QRS}_0},           \\
         &\qquad L(T^{\mathtt{QRS}_1}, T^{\mathtt{QRS}_2}) = [200, 0.9\cdot    
     							      \mathtt{RR0}]\} \\
[l_{53}] &= \{\mathtt{RR1} = t^{\mathtt{QRS}_2}-t^{\mathtt{QRS}_1},           \\
	 &\qquad L(T^{\mathtt{QRS}_1}, T^{\mathtt{QRS}_3}) = 
                               [1.7\cdot\mathtt{RR0}, 2.3\cdot\mathtt{RR0}],  \\
         &\qquad L(T^{\mathtt{QRS}_2}, T^{\mathtt{QRS}_3}) = 
				[1.25\cdot\mathtt{RR1}, 4\cdot\mathtt{RR1}]   \\
	 &\qquad L(T^{\mathtt{QRS}_3}, T^\mathtt{Extrasystole}_e) = [0,0]\}
\end{align*}
\begin{align*}
\Pi_0:& \{t^\mathtt{BeatAnn} = max_t(\psi^2(ECG(t))),
t\in L(T_0,T^\mathtt{BeatAnn}) \}\\
\Pi_1:& \{t^\mathtt{QRS}=max_t(|ECG(t)-mode(ECG)|),
t\in L(T_0,T^\mathtt{QRS}) \}
\end{align*}
}
In the procedures, $T_0$ represents the time origin, in this case the start of
the ECG recording, and therefore $L(T_0,T)$ is the interval of possible
assignments for the variable $T$.

The first pattern $P_0$ is a purely deductive pattern, since it does not
abstract any observable. It is included in the knowledge base in order to allow
the discovery of new beat annotations in those cases in which a rhythm pattern
predicts the presence of an annotation, but it is not in the initial evidence.
The observation procedure $\Pi_0$ uses a wavelet-based method detailed
in~\cite{Martinez04}, performing a wavelet transformation $\psi$ of the signal
in the predicted interval, and emits as hypothesis the time point in which the
energy of the transform is greater.

The $P_1$ pattern abstracts the initial annotations in observations of QRS
complexes. The pattern introduces as temporal constraint $[l_1]$ a maximum
distance between the beat annotation point and the QRS location. The observation
procedure $\Pi_1$ establishes the time instant of the QRS in the point of
maximum deviation of the signal with respect to the baseline, estimated as the
mode of the signal in the observation interval. As shown in
Fig.~\ref{fig:matchrel}, this procedure allows to change the timing of the
initial annotations.

Regular rhythms $q_2$, $q_3$ and $q_4$ share the same pattern grammar $P_2$,
which requires the presence of at least 3 consecutive QRS complexes with a
bounded distance between them. For $\mathtt{NormalRhythm}$, this distance is
established to a heart rate of 45-120 bpm; for $\mathtt{Bradycardia}$, the
admitted rate is 30-60 bpm; and for $\mathtt{Tachycardia}$, 100-300 bpm. The
overlap between patterns allows the correct interpretation of those fragments
that are on the frontier of two rhythms. In addition to these static
constraints, $[l_{2n}]$ introduces an additional constraint limiting the instant
rhythm variation. These patterns do not have an observation procedure $\Pi$,
since the temporal limits are set to the time point of the first and last QRS
complexes.

The extrasystole pattern $P_5$ is also defined only by temporal constraints. An
extrasystole requires the presence of exactly 4 consecutive beats. The first two
define the reference rhythm, while the $[l_{52}]$ constraint sets an upper bound
on the duration of the second RR interval, forcing the third beat to be
premature. $[l_{53}]$ introduces the compensatory pause constraints, requiring
the difference between the second and fourth beats to be approximately two times
the previous RR, and a significant increase in the RR interval between the third
and fourth beats.

With this simple knowledge base, we are able to apply the interpretation
algorithm presented in section~\ref{sec:algorithm} in order to obtain the best
interpretation of a set of beat annotations, which may be obtained by any
external algorithm. To avoid an excessive exploration of the search space if no
interpretations with full coverage factor were found, a restriction in the
search procedure was included, consisting in limiting the size of the
\texttt{open} list to at most $K$ elements (in this case, $K=4$) whenever the
interpretation process exceeds the real duration of the interpreted signal
fragment. This ensures that the interpretation is performed in soft real time.
As a result of the interpretation, a new set of annotations will be generated
from the $\mathtt{QRS}$ observations present in the best interpretation obtained
by the algorithm, provided that the observation is included in a rhythm pattern.

\section{Validation results}
\label{sec:validation}
To validate the ability of the presented framework to revise and correct the
results of classical deductive approaches, we have used a state of the art
algorithm providing a set of beat annotations for an ECG recording. These
annotations are converted to instances of the $\mathtt{BeatAnn}$ observable, and
form the initial observations upon which the abductive interpretation process is
applied. The selected algorithm is the \texttt{WQRS} algorithm~\cite{Zong03},
included in the standard distribution of the WFDB software
package~\cite{Goldberger00}. The validation dataset consists of a selection of
ECG recordings showing regular rhythms and extrasystoles, and comprises all the
18 recordings of 24 hours duration of the Normal Sinus Rhythm (NSR) database and
20 recordings of 30 minutes duration from the MIT-BIH Arrhythmia database, from
the Physionet initiative~\cite{Goldberger00}.

Table~\ref{tab:validation} shows the comparative results of the original
algorithm and the corrected output through abduction, in terms of sensitivity,
positive predictivity, and the combined F1-score. As can be noted, the
abstraction process can slightly decrease the sensitivity, but is compensated
with an improvement in the positive predictivity to get in the majority of cases
an advance in the combined F1-score. Still, cases such as records MIT-103 or
MIT-109 show that the abstraction process is also able to correct sensitivity
failures. In order to prove that the improvement on the detection is
significant, the Wilcoxon statistical test has been applied to the differences
on the F1-score, obtaining a \textit{p-value} of 0.008 on the records of the
Normal Sinus Rhythm database, a \textit{p-value} of 0.033 on the records of the
MIT-BIH Arrhythmia database, and a combined \textit{p-value} of 0.001 on the
full set of records. This demonstrates that the improvement is statistically
significant.

{%
\begin{table}
\footnotesize
\newcommand{\mc}[3]{\multicolumn{#1}{#2}{#3}}
\begin{center}
\begin{tabular}{r|r|r|r|r|r|r|}\cline{2-7}
  & \mc{3}{c|}{\texttt{WQRS}} & \mc{3}{c|}{\texttt{WQRS} + Abduction}\\\hline
\mc{1}{|l|}{\textbf{Record}} & \mc{1}{c|}{\textbf{Se}} & \mc{1}{c|}{\textbf{P+}}
& \mc{1}{c|}{\textbf{F1}} & \mc{1}{c|}{\textbf{Se}} & \mc{1}{c|}{\textbf{P+}} &
\mc{1}{c|}{\textbf{F1}}\\\hline
\mc{1}{|l|}{NSR-16265} & 100.00 & 99.74 & \textbf{99.87} & 99.97 & 99.76 &
	 99.86\\\hline
\mc{1}{|l|}{NSR-16272} & 98.24 & 89.79 &          93.83 & 97.74 & 93.66 &
\textbf{95.66}\\\hline
\mc{1}{|l|}{NSR-16273} & 99.99 & 99.93 &          99.96 & 99.95 & 99.98 &
\textbf{99.96}\\\hline
\mc{1}{|l|}{NSR-16420} & 99.98 & 99.79 &          99.88 & 99.92 & 99.92 &
\textbf{99.92}\\\hline
\mc{1}{|l|}{NSR-16483} & 99.98 & 99.88 &          99.93 & 99.98 & 99.97 &
\textbf{99.97}\\\hline
\mc{1}{|l|}{NSR-16539} & 99.97 & 99.79 & \textbf{99.88} & 99.76 & 99.92 &
         99.84\\\hline
\mc{1}{|l|}{NSR-16773} & 99.99 & 99.96 & \textbf{99.97} & 99.96 & 99.95 &
         99.95\\\hline
\mc{1}{|l|}{NSR-16786} & 100.00 & 99.97 &         99.98 & 100.00 & 99.98 &
\textbf{99.99}\\\hline
\mc{1}{|l|}{NSR-16795} & 99.99 & 99.87 & \textbf{99.93} & 99.96 & 99.88 &
         99.92\\\hline
\mc{1}{|l|}{NSR-17052} & 99.98 & 99.52 &          99.75 & 99.95 & 99.68 &
\textbf{99.81}\\\hline
\mc{1}{|l|}{NSR-17453} & 99.98 & 99.72 &          99.85 & 99.94 & 99.89 &
\textbf{99.91}\\\hline
\mc{1}{|l|}{NSR-18177} & 99.98 & 99.63 &          99.80 & 99.91 & 99.74 &
\textbf{99.82}\\\hline
\mc{1}{|l|}{NSR-18184} & 99.99 & 99.55 &          99.77 & 99.98 & 99.79 &
\textbf{99.88}\\\hline
\mc{1}{|l|}{NSR-19088} & 100.00 & 98.29 &         99.14 & 99.98 & 98.37 &
\textbf{99.17}\\\hline
\mc{1}{|l|}{NSR-19090} & 99.99 & 99.70 &          99.84 & 99.99 & 99.79 &
\textbf{99.89}\\\hline
\mc{1}{|l|}{NSR-19093} & 100.00 & 99.87 & \textbf{99.93} & 99.99 & 99.88 &
\textbf{99.93}\\\hline
\mc{1}{|l|}{NSR-19140} & 100.00 & 99.82 &         99.91 & 100.00 & 99.84 &
\textbf{99.92}\\\hline
\mc{1}{|l|}{NSR-19830} & 99.93 & 98.74 &          99.33 & 99.86 & 98.99 &
\textbf{99.42}\\\hline
\mc{1}{|l|}{\textbf{NSR-Gross}} & \textbf{99.90} & \textbf{99.08} &
\textbf{99.49} & \textbf{99.83} & \textbf{99.40} & \textbf{99.61}\\\hline
\mc{1}{|l|}{MIT-100} & 100.00 & 99.95 & \textbf{99.97} & 99.95 & 99.95 &
         99.95\\\hline
\mc{1}{|l|}{MIT-101} & 99.93 & 99.80 &           99.86 & 99.80 & 100.00 &
\textbf{99.90}\\\hline
\mc{1}{|l|}{MIT-102} & 100.00 & 100.00 & \textbf{100.00} & 100.00 & 100.00 &
\textbf{100.00}\\\hline
\mc{1}{|l|}{MIT-103} & 99.94 & 100.00 &           99.97 & 100.00 & 100.00 &
\textbf{100.00}\\\hline
\mc{1}{|l|}{MIT-104} & 100.00 & 97.58 &           98.78 & 99.25 & 98.93 &
\textbf{99.09}\\\hline
\mc{1}{|l|}{MIT-105} & 99.72 & 91.72 &            95.55 & 98.00 & 97.24 &
\textbf{97.62}\\\hline
\mc{1}{|l|}{MIT-107} & 99.89 & 98.13 &            99.00 & 99.66 & 99.94 &
\textbf{99.80}\\\hline
\mc{1}{|l|}{MIT-108} & 99.59 & 86.20 &            92.41 & 95.54 & 97.79 &
\textbf{96.65}\\\hline
\mc{1}{|l|}{MIT-109} & 99.86 & 100.00 &           99.93 & 100.00 & 100.00 &
\textbf{100.00}\\\hline
\mc{1}{|l|}{MIT-111} & 99.94 & 99.94 & \textbf{99.94} & 99.77 & 100.00 &
          99.88\\\hline
\mc{1}{|l|}{MIT-112} & 100.00 & 99.91 &         99.95 & 100.00 & 100.00 &
\textbf{100.00}\\\hline
\mc{1}{|l|}{MIT-113} & 100.00 & 100.00 & \textbf{100.00} & 100.00 & 99.67 &
          99.83\\\hline
\mc{1}{|l|}{MIT-115} & 100.00 & 100.00 & \textbf{100.00} & 100.00 & 100.00 &
\textbf{100.00}\\\hline
\mc{1}{|l|}{MIT-117} & 100.00 & 100.00 & \textbf{100.00} & 100.00 & 100.00 &
\textbf{100.00}\\\hline
\mc{1}{|l|}{MIT-122} & 100.00 & 99.95 &            99.97 & 100.00 & 100.00 &
\textbf{100.00}\\\hline
\mc{1}{|l|}{MIT-123} & 100.00 & 100.00 & \textbf{100.00} & 99.84 & 99.06 &
          99.45\\\hline
\mc{1}{|l|}{MIT-209} & 100.00 & 99.53 &            99.76 & 100.00 & 99.88 &
\textbf{99.94}\\\hline
\mc{1}{|l|}{MIT-212} & 100.00 & 99.91 &            99.95 & 100.00 & 100.00 &
\textbf{100.00}\\\hline
\mc{1}{|l|}{MIT-230} & 100.00 & 99.95 &            99.97 & 100.00 & 100.00 &
\textbf{100.00}\\\hline
\mc{1}{|l|}{MIT-234} & 100.00 & 100.00 & \textbf{100.00} & 100.00 & 100.00 &
\textbf{100.00}\\\hline
\mc{1}{|l|}{\textbf{MIT-Gross}} & \textbf{99.95} & \textbf{98.58} &
\textbf{99.26} & \textbf{99.62} & \textbf{99.64} & \textbf{99.63}\\\hline
\end{tabular}
\end{center}
\caption{QRS detection correction results}
\label{tab:validation}
\end{table}
}%

\section{Conclusion}
\label{sec:conclusion}
In this paper we have presented an abductive framework to address the problem of
biosignal interpretation. We have conducted a validation study to prove that the
non-monotonic behavior is able to review the results of classical approaches,
trying to enhance a state of the art QRS detector using a very simple knowledge
base. The objective was not to provide a new algorithm for QRS detection, but to
prove that an abductive approach can overcome the limitations of deductive
methods. Results demonstrate the capacity of our proposal to correct and improve
the outcomes of the original algorithm. Future plans include building a more
complete knowledge base, modeling other ECG constituents, like P and T waves,
explicitly considering the morphology of the significant waves, and including
more rhythm patterns. This is expected to provide a more interpretative
information, helping to improve previous results on ECG processing at all
abstraction levels, from QRS detection to arrhythmia recognition.  


\section*{Acknowledgment}
T. Teijeiro is funded by an FPU grant from the Spanish Ministry of Education
(MEC) (ref. AP2010-1012).



%
\bibliography{bibliography}
\bibliographystyle{IEEEtran}

\section*{Copyright Disclaimer}
\copyright 2014 IEEE. Personal use of this material is permitted. Permission
from IEEE must be obtained for all other uses, in any current or future media,
including reprinting/republishing this material for advertising or promotional
purposes, creating new collective works, for resale or redistribution to servers
or lists, or reuse of any copyrighted component of this work in other works.

\end{document}